\def\year{2022}\relax
\documentclass[letterpaper]{article} 
\usepackage{aaai22}  
\usepackage{times}  
\usepackage{helvet}  
\usepackage{courier}  
\usepackage[hyphens]{url}  
\usepackage{graphicx} 
\usepackage{natbib}  
\usepackage{caption} 
\DeclareCaptionStyle{ruled}{labelfont=normalfont,labelsep=colon,strut=off} 
\usepackage{algorithm}
\usepackage{algorithmic}

\usepackage{amsmath}
\usepackage{amssymb}
\usepackage{multirow,balance}

\usepackage{hyperref}
\usepackage{subfigure}

\usepackage{newfloat}
\usepackage{listings}
\floatstyle{ruled}
\newfloat{listing}{tb}{lst}{}
\floatname{listing}{Listing}
\title{Miti-DETR: Object Detection based on Transformers with Mitigatory Self-Attention Convergence}
\author {
    Wenchi Ma\textsuperscript{\rm 1},
    Tianxiao Zhang\textsuperscript{\rm 2},
    Guanghui Wang\textsuperscript{\rm 3}\footnote{This work was partly supported by the Natural Sciences and Engineering Research Council of Canada (NSERC) under grant no. RGPIN-2021-04244, and the United States Department of Agriculture (USDA) under Grant no. 2019-67021-28996.}
}
\affiliations {
    \textsuperscript{\rm 1} Machine Learning Group, Rekor Systems, Inc., Columbia, MD, USA, 21046\\
    \textsuperscript{\rm 2} Department of Electrical Engineering and Computer Science, University of Kansas, Lawrence, KS, USA, 66045\\
    \textsuperscript{\rm 3} Department of Computer Science, Ryerson University, Toronto, ON Canada, M5B 2K3\\
}

\usepackage{bibentry}

\begin{document}

\maketitle

\begin{abstract}
Object Detection with Transformers (DETR) and related works reach or even surpass the highly-optimized Faster-RCNN baseline with self-attention network architectures. Inspired by the evidence that pure self-attention possesses a strong inductive bias that leads to the transformer losing the expressive power with respect to network depth, we propose a transformer architecture with a mitigatory self-attention mechanism by applying possible direct mapping connections in the transformer architecture to mitigate the rank collapse so as to counteract feature expression loss and enhance the model performance. We apply this proposal in object detection tasks and develop a model named Miti-DETR. Miti-DETR reserves the inputs of each single attention layer to the outputs of that layer so that the ``non-attention" information has participated in any attention propagation. The formed residual self-attention network addresses two critical issues: (1) stop the self-attention networks from degenerating to rank-1 to the maximized degree; and (2) further diversify the path distribution of parameter update so that easier attention learning is expected. Miti-DETR significantly enhances the average detection precision and convergence speed towards existing DETR-based models on the challenging COCO object detection dataset. Moreover, the proposed transformer with the residual self-attention network can be easily generalized or plugged in other related task models without specific customization.
\end{abstract}

\section{Introduction}
\noindent The attention mechanism has been effectively used in transformer networks ~\cite{vaswani2017attention}, not only in the application of long-range sequential knowledge, such as natural language processing~\cite{devlin2018bert}, speech recognition ~\cite{luo2021simplified}, but also in computer vision tasks~\cite{gajurel2021fine, carion2020end, dai2021up, zhu2020deformable, sajid2021parallel}, where DETR has achieved competitive performance as an end-to-end object detector~\cite{carion2020end}. Attention mechanism, transformer networks and DETR, thus, have become the research focuses, where the inner workings of transformers and attention, the training and optimization challenge of DETR, etc., have been regarded shedding light for future works. 

The attention-mechanism based transformer networks realize the most generalized deep learning model in terms of computer vision and image tasks. In transformer networks, one pixel in an image cares about all the other pixels in that image so that any single region obtains and integrates relevance with all other regions, which is in comparison with CNN that any single pixel cares about its immediate neighborhood and then what the neighborhood as a whole cares about is its immediate neighborhood. This could be a good explanation of why DETR can be on par with state-of-the-art classifiers in terms of classification accuracy. It also explains the strong inductive bias of self-attention. While the research in~\cite{dong2021attention} demonstrates that pure self-attention networks (SANs) would lead to the loss of expressive power doubly exponentially with respect to network depth, and the output converges with a cubic rate to a rank one matrix that has identical rows. 

\begin{center}
\begin{figure*}
    \includegraphics[width=1.0\linewidth]{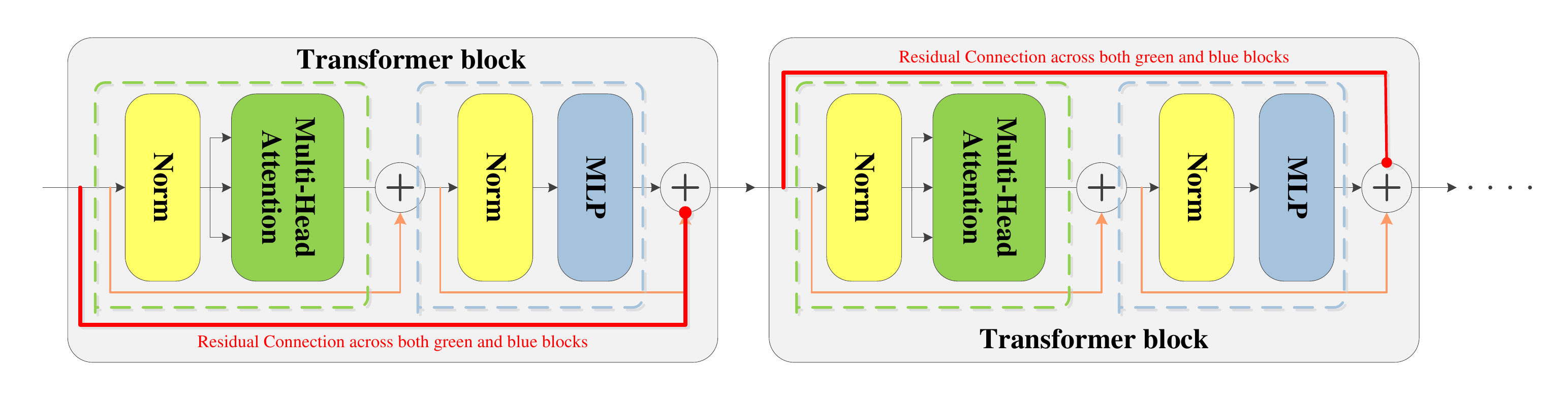}\hfill
    \vspace{-2mm}
    \caption{Residual Attention Network in Transformer}
	\label{fig:residual}
 	\vspace{-3mm}
\end{figure*}
\end{center}

Inspired by the analysis that skips connections play a key role in mitigating rank collapse in transformers, we propose the Miti-DETR detector model where residual self-attention network architecture is introduced. Specifically, the inputs of a multi-head self-attention layer are short connected to its outputs, as shown in Figure~\ref{fig:residual}. This connection skips the Multi-layer perceptions (MLP), which is usually considered rendering the model less sensitive to its input perturbations~\cite{cranko2018lipschitz}. Thus, the relative "non-attention" features are integrated to avoid the outputs dramatically converging to rank one matrix. In summary, this work makes two critical contributions to the current DETR models:
\begin{itemize}
    \item The training and optimization challenges of DETR could be better solved based on the network model itself, which is data-independent and could be easily applied in the related models. It is verified that the proposed model could significantly speed up the training convergence so as to avoid the extremely long time training schedule. 
    \item The proposed model can further enhance the object detection performance of DETR models. From the perspective of models themselves, we address the limitations of DETR models where the transformer networks tend to lose efficient feature expression power by proposing the residual self-attention network. Thus, the proposed Miti-DETR could better stop the outputs from degeneration and achieve nearly 3\% higher performance than the original DETR models.
\end{itemize}

We evaluate Miti-DETR on one of the most popular and challenging object detection datasets, COCO, and compare its performance with the traditional DETR-related models. Our experiments show that our model is capable of further enhancing the DETR models' performance. Specifically, Miti-DETR brings the superiority of DETR detecting large objects into full play, enabled by the protection of global expression power. Code is available on \url{https://github.com/wenchima/Miti-DETR}.

\section{Related Work}
This work is built on prior researches of attention mechanism~\cite{vaswani2017attention}, feature mapping and propagation~\cite{he2016deep} and object detection with transformer~\cite{carion2020end}.

\subsection{Attention Mechanism}
Attention-based architectures have become ubiquitous in machine learning, which brings about better learning for long-sequence and large-range knowledge~\cite{bahdanau2014neural}~\cite{ramachandran2019stand}. They have permeated machine learning applications across data domains, such as natural language processing~\cite{devlin2018bert}, speech recognition~\cite{luo2021simplified}, and computer vision~\cite{bello2019attention}~\cite{carion2020end}\cite{sajid2021audio}.  Attention mechanisms are neural network layers that aggregate information from the entire input sequence~\cite{bahdanau2014neural}. They allow modeling of dependencies without regard to their distance in the input or output sequences~\cite{bahdanau2014neural}~\cite{kim2017structured} and the early such attention mechanisms models mostly are applied in conjunction with the recurrent network~\cite{parikh2016decomposable}. 

Self-attention is an attention mechanism that relates different positions in a single sequence so as to compute a sequence representation~\cite{cheng2016long}~\cite{lin2017structured}. End-to-end memory networks are based on a recurrent attention mechanism rather than sequence aligned recurrence, which shows advantages on simple-language question answering and language modeling tasks~\cite{sukhbaatar2015end}. Transformer, currently, is the first transduction model which entirely relies on self-attention to compute representations of its input and output without using sequence aligned RNNs or convolution~\cite{vaswani2017attention}. They introduce self-attention layers to Non-Local Neural Networks~\cite{wang2018non}. One advantage of attention-based models is the global computations and superior memory, making them more suitable compared with RNNs on long sequences. Transformers now have shown its replacing role 
then RNNs in many problems in natural language processing, speech processing and computer vision~\cite{parmar2018image}~\cite{synnaeve2019end}. 

Recently, researchers find that pure self-attention networks (SANs), for example, transformers with skip connections and multi-layer perceptrons (MLPs) disabled, lose expressive power doubly exponentially with respect to network depth. They prove that the output converges with a cubic rate to a rank one matrix with identical rows~\cite{dong2021attention}. Their analysis verifies that skip connections are the key in mitigating rank collapse, and MLPs can slow down the convergence by increasing their Lipschitz constant. This research inspires our deep thoughts towards the current object detection models with transformer. We try to excavate the inner specialty of the transformer so as to solve the existing problem, especially in object detection applications.

\subsection{Object Detection with Transformer}
Previously, the mainstream object detection models make predictions relative to some initial guesses \cite{ma2021semantic}. Two-stage detectors~\cite{ren2015faster, zhang2020efficient} predict bounding boxes relative to proposals, and single-stage methods make predictions based on anchors~\cite{ma2020mdfn, ma2020location} or other features~\cite{li2021sgnet, zhang2021six}. The performance of these models heavily depends on the set of initial guesses. Anchor-free detection technologies assign positive and negative samples to feature maps by a grid of object centers~\cite{law2018cornernet}. 

DETR is recently proposed that successfully apply transformer in object detection that is conceptually simpler without handcrafted process by direct set prediction~\cite{carion2020end}.  DETR utilizes a simple architecture, by combining convolutional neural networks (CNNs) and Transformer encoder-decoders. Deformable DETR is proposed to improve the problems of slow-convergence and limited feature spatial resolution by making its attention modules only attend to a small set of key points around a reference~\cite{zhu2020deformable}. UP-DETR is inspired by the pre-training transformers in natural language processing and proposes the random query patch detection to unsupervisedly pre-train the transformer of DETR~\cite{dai2021up}, which boosts the performance significantly. While these two models both solve the problem from the data end, one by ImageNet pre-training, the other through multi-scale feature representation. Compared with the previous works, Miti-DETR tries to solve the existing problems from the inner property of transformer so that it is data-independent and more practical, which is meaningful as an improvement towards DETR. 

\section{Miti-DETR}
\begin{center}
\begin{figure*}
    \includegraphics[width=1.0\linewidth]{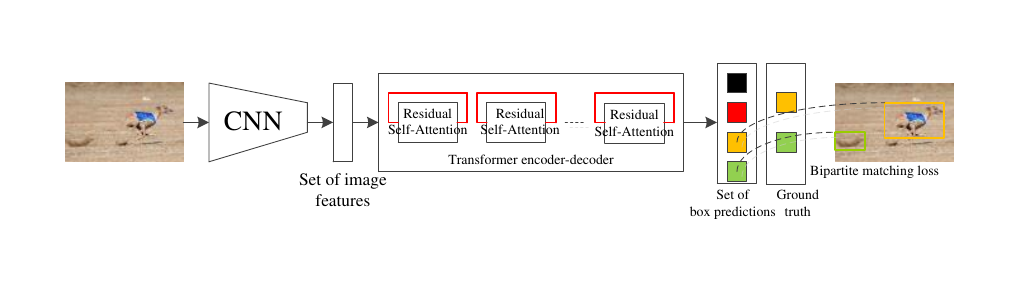}\hfill
    \vspace{-8mm}
    \caption{Architecture Streamline of Miti-DETR}
	\label{fig:overview}
\end{figure*}
\end{center}

The proposed Miti-DETR model maintains the simple architecture as the original DETR, as depicted in Figure~\ref{fig:overview}. It contains three main components: a CNN working as the feature extractor; the transformer encoder and decoder with the proposed residual self-attention network; and the final feed-forward networks (FFN) working for the object detection prediction. We adopt the effective bipartite matching loss for direct prediction~\cite{carion2020end}. In this section, we mainly discuss why we need to introduce the proposed residual connection, and how to build up the corresponding transformer architecture,  and finally, we show the proof of why this new architecture could bring about mitigatory self-attention convergence. 
\subsection{Attention Network Loses Rank}
The attention mechanism has become ubiquitous by its outstanding performance of learning long-range knowledge both in timing and spatial sequence~\cite{bahdanau2014neural}~\cite{vaswani2017attention}. However, it has been certified that the pure-attention networks (SANs), by disabling the skip connections and multi-layer perceptions (MLPs), tends to losing expressive power with the network getting deep, leading to the output converging to a rank one matrix with cubic rate ~\cite{dong2021attention}. This can be expressed as below~\cite{dong2021attention}.
\begin{equation}\label{equation_singlehead}
\lVert res(SAN(\textbf{\textit{X}}))\rVert_{1, \infty} \le \left(\frac{4\alpha}{\sqrt{d_{qk}}}\right)^{\frac{3^L-1}{2}}\lVert res(\textbf{\textit{X}})\rVert_{1, \infty}^{3^{L}}
 \end{equation}
which agrees to a double exponential rate of convergence. 
The residual item in the above expression works as, 

\begin{equation} \label{equation_res}
res(\textbf{\textit{X}}) = \textbf{\textit{X}} - 1\textbf{x}^{T}, where\ \textbf{x} = argmin_{\textbf{x}}\lVert \textbf{\textit{X}} - 1\textbf{x}^{T}\rVert  
\end{equation}
Here the input $\textbf{\textit{X}}$ is a $n \times d_{in}$ matrix that includes $n$ tokens. $\textbf{\textit{W}}_{QK}^{l}$ and $\textbf{\textit{W}}_{V}^{l}$ are the corresponding value weight matrices. It is noted that the bound in Equation~\ref{equation_singlehead} ensures $\lVert res(SAN(\textbf{\textit{X}}))\rVert_{1, \infty}$ converge for all the small residual's inputs whenever $4\alpha \le \sqrt{d_{qk}}$. The detailed proofs can be found in~\cite{dong2021attention}.



\subsection{Skip Connection and MLP Counteract Degeneration}
There is a natural but pertinent question: If the (pure) attention network degenerates to a rank one matrix with the increase of depth, why do attention-based transformer networks work in applications? It is verified that the presence of skip connections is crucial that prevents the SAN from degenerating to rank one and the Multi-layer perceptrons (MLPs) help control the convergence rate~\cite{dong2021attention}. This argument was originally proved by the discussion of the bounds for the residual. 


We denote the output expression of an MLP with depth $L$ and width $H$ as
\begin{equation}
\textbf{\textit{X}}^{l+1} = f_{l}\left(\sum_{h\in[H]} \textbf{\textit{P}}_{h}\textbf{\textit{X}}^{l} \textbf{\textit{W}}_{h} \right),
\end{equation}
where $P_{h}$ is the $n \times n$ row-stochastic matrix. $W_{h}$ is the weights matrix. 

$\lambda_{l,1,\infty}$ here is used to represent the Lipschitz constant of $f_{l}$ concerning $l_{1,\infty}$ norm. The upper bound for the residual is derived in the following ~\cite{dong2021attention}:
\begin{equation} \label{upper_bound}
\lVert res(\textbf{\textit{X}}^{L})\rVert_{1, \infty} \le \left(\frac{4\alpha H\lambda}{\sqrt{d_{qk}}}\right)^{\frac{3^L-1}{2}}\lVert res(\textbf{\textit{X}})\rVert_{1, \infty}^{3^{L}}    
\end{equation}
which agrees with a double exponential rate of convergence.

Thus, the convergence rate can be adjusted by the MLPs' Lipschitz constants $\lambda_{f,1,\infty}$, which certifies that more powerful MLPs bring about slower convergence. This shows the hard struggle between self-attention layers and the MLPs whose nonlinearity contributes to increasing the rank~\cite{dong2021attention}.

\subsection{Residual Self-attention Network}
By observing the current transformer network, we find that the skip connections are across ``independent modules". Here the ``independent modules" refer to the multi-head self-attention network and the feed-forward fully connection network in transformer encoder and decoder layer architectures. Based on the analysis towards the functions of SANs, skip-connection and MLPs, we can define the features' attention levels and provide the sequence. Considering one single transformer layer, we use $\textbf{\textit{X}}^{l}$ to denote its inputs, $SAN(\textbf{\textit{X}}^{l})$ to denote the outputs of SAN network, and $\textbf{\textit{X}}^{l+1}$ as the outputs after MLPs. If $g$ is defined as the mapping function of features' attention level, we can derive the following conclusion.
\begin{equation}
g(SAN(\textbf{\textit{X}}^{l})) \geq g(\textbf{\textit{X}}^{l+1}) \geq g(\textbf{\textit{X}}^{l})
\end{equation}

Combined with the advantage brought by skip connections, we believe that the skip connections that dramatically diversify the path distribution are the structural factor that explains the degeneration counteraction. This structural factor brings in the diversity of attention levels. Essentially, we believe that it is the diversity of features' attention levels that mitigate the strong inductive bias towards ``token uniformity" of the self-attention networks. Inspired by this idea, we propose the residual self-attention network, in which we reserve the inputs of each single attention layer to the outputs of this layer so that the "non-attention" information could participate in the feature attention propagation. Thus, the diversity of feature attention levels is maximized. The corresponding architecture is depicted in Figure~\ref{fig:residual}. 
In the sequence, we provide proof for the convergence property of the new self-attention network with the proposed residual connection. 

The output of the $l_{th}$ layer attention network, in this case, can be expressed as
\begin{equation}
\textbf{\textit{X}}^{l+1} = f_{l}\left(\sum_{h\in[H]} \textbf{\textit{P}}_{h}\textbf{\textit{X}}^{l} \textbf{\textit{W}}_{h} \right) + \textbf{\textit{X}}^{l},
\end{equation}
then the $l_{th}$ layer output not considering the residual connection is,
\begin{equation}
\textbf{\textit{X}}^{l+1} - \textbf{\textit{X}}^{l} = f_{l}\left(\sum_{h\in[H]} \textbf{\textit{P}}_{h}\textbf{\textit{X}}^{l} \textbf{\textit{W}}_{h} \right).
\end{equation}

We follow the definition of the residual in Equation~\ref{equation_res}, 
\begin{equation}
res(\textbf{\textit{X}}^{l+1} - \textbf{\textit{X}}^{l}) = (\textbf{\textit{X}}^{l+1} - \textbf{\textit{X}}^{l}) - 1\textbf{x}^{T}, 
\end{equation}
where\
\begin{equation}
\textbf{x} = argmin_{\textbf{x}}\lVert (\textbf{\textit{X}}^{l+1} - \textbf{\textit{X}}^{l}) - 1\textbf{x}^{T}\rVert, 
\end{equation}
while the proposed residual connection skips both the multi-head attention layer and the MLPs, the actual output residual should be
\begin{equation}
res(\textbf{\textit{X}}^{l+1}) = \textbf{\textit{X}}^{l+1} - 1\textbf{x}^{T}, 
\end{equation}
where\ 
\begin{equation}
\textbf{x} = argmin_{\textbf{x}}\lVert (\textbf{\textit{X}}^{l+1} - \textbf{\textit{X}}^{l}) - 1\textbf{x}^{T}\rVert.
\end{equation}

Thus, $res(\textbf{\textit{X}}^{l+1})$ can also be expressed in the following equation
\begin{equation}
res(\textbf{\textit{X}}^{l+1}) = \textbf{\textit{X}}^{l} + \epsilon, 
\end{equation}
where
\begin{equation}
\ \epsilon < \textbf{\textit{X}}^{l} and \epsilon = argmin_{\epsilon} \lVert \textbf{\textit{X}}^{l+1} - \textbf{\textit{X}}^{l} - \epsilon\ \rVert. 
\end{equation}

With the introduced residual connection, $res(\textbf{\textit{X}}^{l+1})$ anchors at the inputs $\textbf{\textit{X}}^{l}$ of the current layer. Its fluctuation depends on the perturbance $\epsilon$ of the inputs and the original convergence performance of the attention network. 
We compare the above three equations, $res(\textbf{\textit{X}}^{l+1} - \textbf{\textit{X}}^{l})$ still satisfies the Equation~\ref{upper_bound} by a convergence upper bound, and
\begin{equation}
\lVert res(\textbf{\textit{X}}^{l+1}) \rVert > \lVert res(\textbf{\textit{X}}^{l+1} - \textbf{\textit{X}}^{l}) \rVert.
\end{equation}

Thus, the proposed residual self-attention network brings about an anchor for the convergence of each attention layer, which propagates through all attention layers in the transformer architecture and helps render the model keep sensitive to the input perturbation, which further slows down the convergence and counteracts the rank collapse. This research conclusion is verified by concrete learning performance in the experiment section. 

\begin{figure*}[t]
    \subfigure{\includegraphics[width=0.33\linewidth]{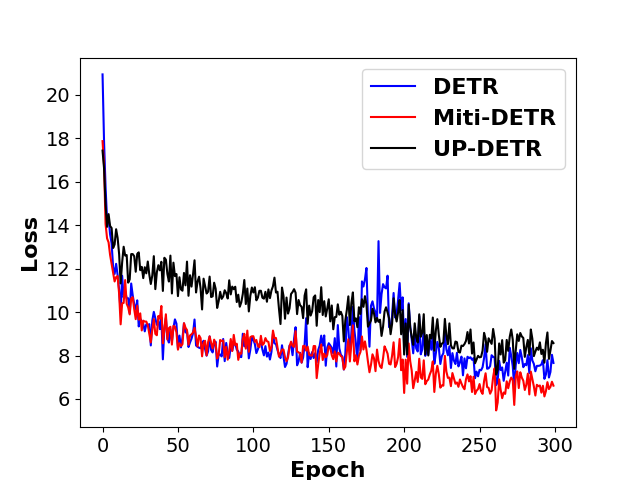}}
    \subfigure{\includegraphics[width=0.33\linewidth]{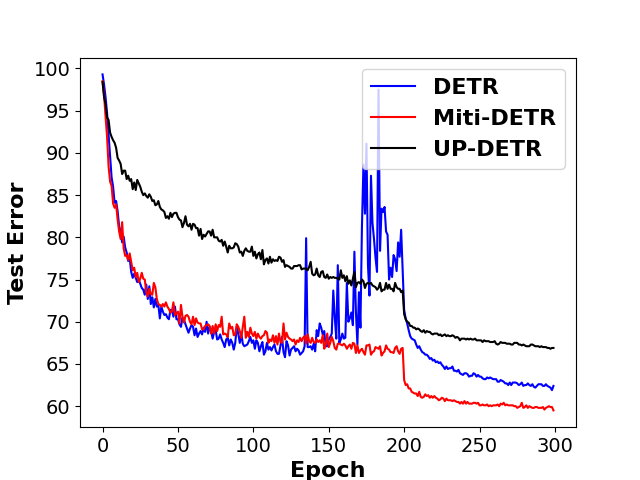}}
    \subfigure{\includegraphics[width=0.33\linewidth]{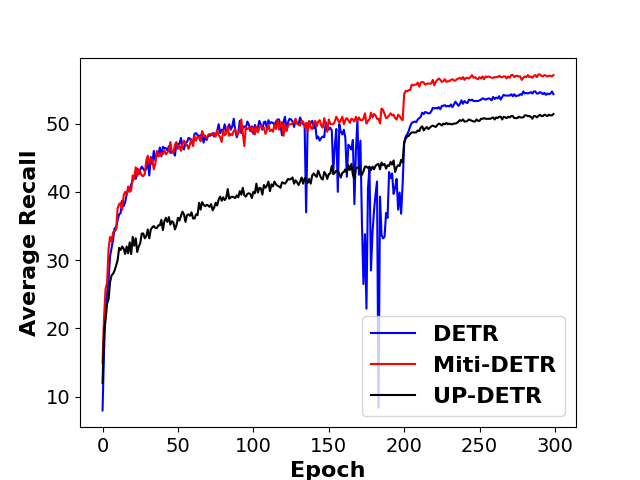}}
    \vspace{2mm}
    \caption{\footnotesize Comparison of the learning process distributions on COCO: {\bf (left to right)} Learning Loss, Test Error, and Average Recall.}
	\label{fig:learning curves}
	\vspace{3mm}
\end{figure*}

\subsection{Transformer with Mitigatory Convergence in Object Detection}
Based on the above analysis, the proposed residual attention architecture enables the transformer to have mitigatory convergence performance theoretically. We follow the same architecture streamline as the DETR model but apply our proposed transformer network as the corresponding encoder-decoder transformer. Specifically, Midi-DETR consists of a convolutional CNN backbone, self-encoder and decoder attention network~\cite{carion2020end} and prediction feed-forward network (FFNs). 

In this work, we design the residual self-attention network in every transformer encoder and decoder layer, as the illustration shows in the transformer block in Figure 2. The residual connection bridges the inputs and outputs of a single transformer layer in a short connection way. This path bypasses the composite module of multi-head self-attention network and the feed-forward fully connection network, forming the new ``non-attention" feature propagation. Then we concatenate the original inputs of the current layer and the outputs from the MLPs and normalization layer at the top of the transformer layer. Thus, this proposed new transformer network works as an independent module that can be implemented in any deep learning framework that provides a common CNN backbone and a transformer architecture implementation. 

\section{Experiments}
We show that Miti-DETR achieves competitive results compared to the original DETR and UP-DETR in quantitative evaluation on COCO~\cite{lin2014microsoft}. Then we provide a detailed analysis towards the training and learning progress, with insights and qualitative results. Then, we provide the detection accuracy results of Miti-DETR on COCO and shows its leading performance at multiple measuring criterion. To show the advantage of speeding up convergence and effective optimization, we compare Miti-DETR and UP-DETR~\cite{dai2021up} and present the detection results and analysis as well. The corresponding experimental settings and implementation details come below. 

\setlength{\tabcolsep}{4pt}
\begin{table*}
\begin{center}
\begin{tabular}{|c|c|c|c|c|c|c|c|c|}
\hline
Model & Backbone & \#Epoch & AP & AP$_{50}$ & AP$_{75}$ & AP$_{S}$ & AP$_{M}$ & AP$_{L}$\\
\hline
DETR & R50 & 300 & 37.6 & 57.8 & 39.3 & 18.0 & 40.6 & 55.7 \\
UP-DETR & R50 & 300 & 33.1 & 50.9 & 34.2 & 14.6 & 35.0 & 50.1 \\
Miti-DETR & R50 & 300 & 40.5 & 60.4 & 42.7 & 19.7 & 43.9 & 59.3\\
\hline
\end{tabular}
\end{center}
\caption{Detection accuracy on COCO.}
\label{table:coco}
\end{table*}
\setlength{\tabcolsep}{1.4pt}

\setlength{\tabcolsep}{4pt}
\begin{table*}
\begin{center}
\begin{tabular}{|c|c|c|c|c|}
\hline
Model & Backbone & Avg Eval Time (s) & \#Params & Accuracy (AP)\\
\hline
DETR & R50 & 484 & 41302368 & 37.7 \\
UP-DETR & R50 & 496 & 41302880 & 33.1 \\
Miti-DETR & R50 & 486 & 41302368 & 40.5 \\
\hline
\end{tabular}
\end{center}
\caption{General performance comparison in terms of efficiency and accuracy}
\label{table:tradeoff}
\end{table*}
\setlength{\tabcolsep}{1.4pt}

\subsection{Implementation Details}
We train the related models in the work with AdamW~\cite{loshchilov2017decoupled}. The transformer weights are initialized with Xavier init~\cite{glorot2010understanding} and the backbone is the ImageNet-pretrained ResNet model~\cite{he2016deep}. In this work, we report the results with the backbone of ResNet-50, which is a relatively basal option. All the other hyperparameters in this experiment strictly follow the setting of DETR~\cite{carion2020end}. 

We use the training schedule of 300 epochs with a learning rate drop of 10 after 200 epochs. All the training images are passed over the model for a single epoch. We train the related models on 4 P100 GPUs, which means each GPU processes two images at the same time in this setting. We apply the pretrained backbone network of DETR R50 provided by DETR official code. \footnote{https://github.com/facebookresearch/detr}

\subsection{Dataset}
We conduct the experiments on the COCO2017 detection dataset~\cite{lin2014microsoft}, which contains 118K training images and 5K validation images. On average, there are 7 instances, up to 63 instances in a single image in the training set, including small and large objects in the same images~\cite{carion2020end}. We report Average Precision (AP) as bounding box AP, under the integral metric with multiple thresholds. For the comparison with state-of-the-art models, we report the validation AP from the highest epoch performance.  

\begin{center}
\begin{figure*}
    \includegraphics[width=1.0\linewidth]{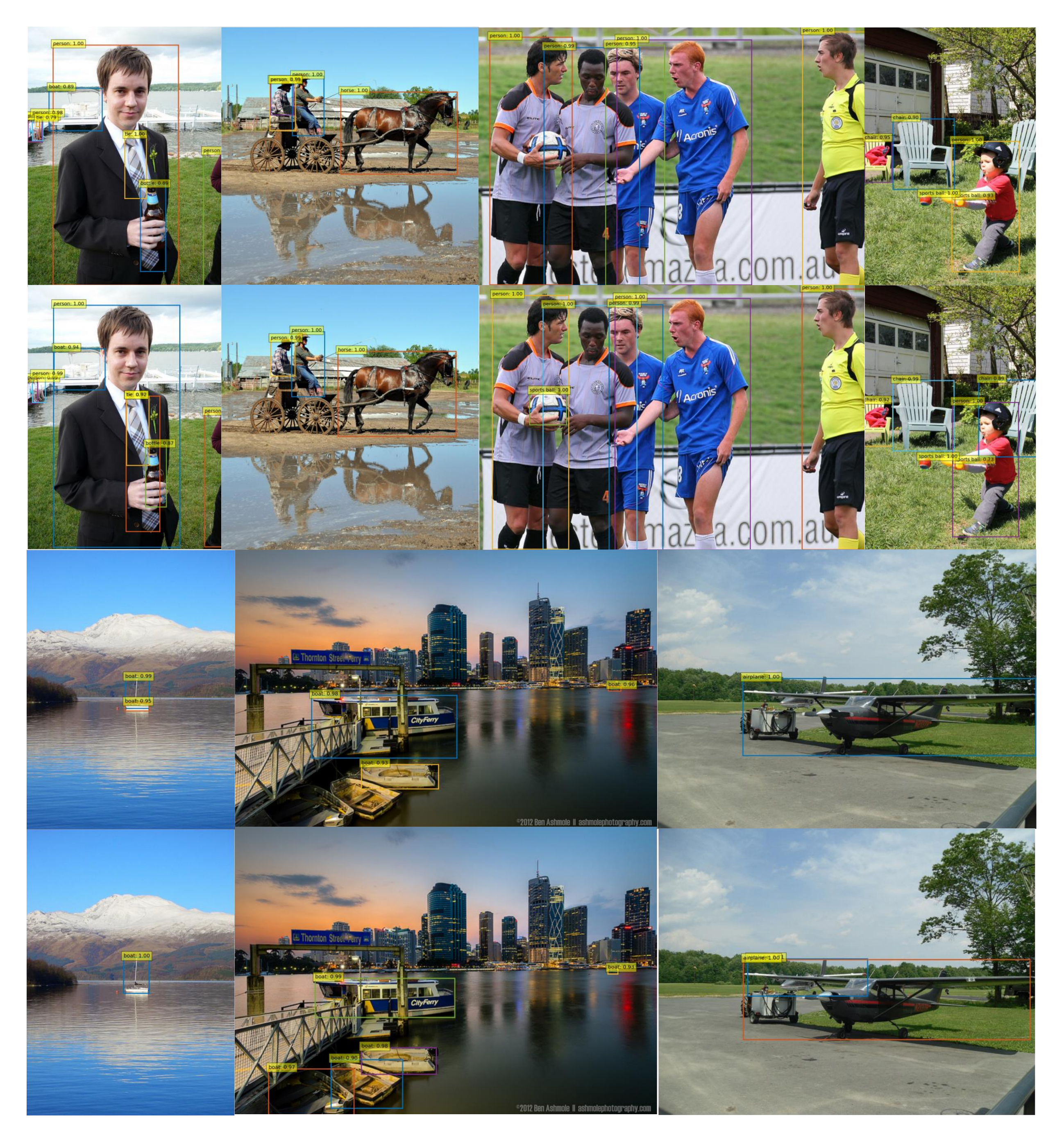}\hfill
    \caption{Visualization Detection Results Comparison. Images in the first row and the second row are the results from DETR and Miti-DETR respectively.}
	\label{fig:bbox}
\end{figure*}
\end{center}

\subsection{Training and Learning}
Typically, transformers are trained with Adam or Adagrad optimizers with very long training schedules and dropout, which is true for DETR models as well. Despite this disadvantage, both UP-DETR and Miti-DETR are designed to make a transformer-based detector with faster convergence. The difference is that UP-DETR focuses on solving the problem by pre-training the transformer network, while Miti-DETR works on handling this problem by attention mechanism and transformer network itself. We report the training loss curves and test error curves of all the experimental models during the learning process. 

\subsection{Detection Results}

\textbf{Setup.} The models in the experiment settings are fine-tuned on COCO train2017 (approximate of 118k images) and evaluated on val2017.  A comprehensive comparison, including AP, AP$_{50}$, AP$_{75}$, AP$_{S}$, AP$_{M}$ and AP$_{L}$, is reported. Moreover, we also show the curves of training loss, test error, and average recall in the learning process and make a comparison among the related models. In order to measure the general performance of the related models considering the trade-off of model size, efficiency and performance, we present the quantitative results of the corresponding properties, including average evaluation time (averaged on 10 times' evaluation) and the number of model parameters. 

\noindent \textbf{Results.} In Figure~\ref{fig:learning curves}, Miti-DETR outperforms DETR for the entire learning process in terms of the convergence speed, showing a clear advantage, especially after the 150 epoch schedule. The statistic of test error in the middle figure in Figure~\ref{fig:learning curves} is the remaining value after deducting Average Precision (AP) from 1 at each epoch. Moreover, DETR shows an unstable learning state, even divergence, between epoch 150 and epoch 200, the end of the first learning rate schedule. While both UP-DETR and Miti-DETR appear to have very stable convergence procedures during the entire process and all of the loss, test error and recall curves show a consistent trend. This is in accord with the theory that the original transformer tends to rank collapse. It is noted that even if both UP-DETR and Miti-DETR seem to be helpful for stabilizing the convergence of DETR, Miti-DETR has a sharply faster convergence performance. 

We can also see from the results that although DETR returns to the normal track of fast convergence after the 200 epoch schedule, the unstable state could be the reason that lowers down the final converging performance of DETR. After the learning rate is reduced at epoch 200, the Miti-DETR averagely keeps the 0.9 test error lower than that of DETR. A similar situation applies to the average recall curves. This experiment suggests the proposed residual self-attention network could effectively improve the convergence of DETR, where the unstable learning procedure caused by the rank-1 trend is avoided by the residual connection across composite network modules in transformer, basically in an end-to-end method without further pre-processing. 

Table~\ref{table:coco} shows the AP statistic results of DETR, UP-DETR and the proposed Miti-DETR. As shown in the table, Miti-DETR generally leads DETR by 3\% in terms of AP. For AP$_{75}$, the highest threshold, Miti-DETR even surpasses DETR by nearly 4\%. This indicates that Miti-DETR generally prominently improves the detection quality of DETR. More detected bounding boxes have higher IoU with the ground truth. Considering the size property of objects, Miti-DETR yields significant advantages over DETR, ranging from small, medium, and large thresholds. Although Miti-DETR has no more than 2\% higher than DETR in terms of AP$_{S}$, this is still a big enhancement considering small objects are hard to handle for current object detectors, and it's an existing problem of DETR as well. As for UP-DETR, it could not obtain better results under our experimental settings, where the limited number of epoch and GPU bring about insufficient training compared with the experiments in the UP-DETR paper~\cite{dai2021up}. From this perspective, the proposed Miti-DETR is more robust in different experimental settings. Combined with Figure~\ref{fig:learning curves}, it is worth noting that the stable learning process effectively contributes to the final detection accuracy of DETR and the Miti-DETR shows a promising research direction for future research based on the transformer network itself. 

Figure~\ref{fig:bbox} presents the visualization detection results in the comparison between the DETR and Miti-DETR. It can be seen that Miti-DETR has less false detection than DETR, such as the tie in the first set of images and the coachman in the second group of images. On the other hand, Miti-DETR seems to have less leak detection, such as the boats and the aircraft in the last two groups of images, respectively.

We also provide the superior performance comparison of Miti-DETR in terms of the trade-off of the model size, running speed and detection performance in Table~\ref{table:tradeoff}. The statistic of evaluation time is based on the entire COCO evaluation dataset, calculating the processing time on the 5K images. Miti-DETR keeps the same model size with the same number of parameters and nearly the same inference time. While for UP-DETR, it increases the model size and lowers down the running speed. Thus, we can conclude that Miti-DETR shows a better practical significance by solving the problem based on the model itself. 


\section{Conclusion}
We have presented Miti-DETR, a DETR object detector based on the transformer with mitigatory self-attention convergence. The model outperforms DETR by a large advantage. The proposed core technique, the residual self-attention network, is verified capable of preventing the attention network from losing rank based on the transformer network itself. Miti-DETR is straightforward to implement and has a flexible structure where the residual self-attention network could be extensible to other attention-mechanism models or tasks. In addition, it achieves significantly better performance on small objects than DETR, thanks to the effective processing of global information and the stable convergence procedure. This work could inspire further research towards the working principles of attention mechanism and transformer network, especially for the effective and efficient processing towards the global attention. Accordingly, current models could be more productive with the training process and a more comprehensive method of combining global information and local features based on the transformer network itself is expected. 

\balance



\balance
\end{document}